\documentclass[10pt,twocolumn,letterpaper]{article}

\usepackage{cvpr}
\usepackage{times}
\usepackage[pdftex,colorlinks]{hyperref}
\usepackage[numbered]{bookmark}
\usepackage{tabularx}
\usepackage{amsmath,amssymb} 
\usepackage{multirow}
\usepackage{varwidth}
\usepackage{color}
\usepackage{listings}
\usepackage{enumitem}
\usepackage{graphicx}
\usepackage{subfigure}
\usepackage{caption}
\usepackage[Algorithm]{algorithm}
\usepackage{algpseudocode}
\usepackage{booktabs} 
\usepackage{xcolor,colortbl}

\cvprfinalcopy 
\def\cvprPaperID{37} 

\begin{document}

\title{Zoom Out-and-In Network with Recursive Training for Object Proposal}

\author{Hongyang Li\textsuperscript{1}, Yu Liu\textsuperscript{2}, Wanli Ouyang\textsuperscript{1} and  Xiaogang Wang\textsuperscript{1} \vspace{.3cm}  
	\\
	{
\textsuperscript{1} The Chinese University of Hong Kong  ~~~~\textsuperscript{2} SenseTime Group Ltd. }\\
{\tt\small yangli@ee.cuhk.edu.hk~~liuyu@sensetime.com~~\{wlouyang,xgwang\}@ee.cuhk.edu.hk
}}

\maketitle

\begin{abstract}



In this paper, we propose a zoom-out-and-in  network for generating object proposals. 
We utilize different resolutions of feature maps in the network to detect object instances of various sizes.
Specifically, we divide the anchor candidates into three clusters based on the scale size and place them
on feature maps of distinct strides to detect small, medium and large objects, respectively.
Deeper feature maps contain region-level semantics which can help shallow counterparts to identify small objects.
Therefore we design a zoom-in sub-network to increase the resolution of 
high level features via a deconvolution operation. 
The high-level features with high resolution are then combined and merged with low-level features to detect objects.
Furthermore, we devise a recursive training pipeline to 
consecutively regress region proposals at the training stage in order to match the iterative regression at the testing stage.
We demonstrate the effectiveness of the proposed method on ILSVRC DET and MS COCO datasets, where our algorithm 
performs better than the state-of-the-arts in various evaluation metrics. 
It also increases average precision by around 2\% in the detection system.

\end{abstract}

\section{Introduction}
\label{Sec:intro}
%

\begin{figure}[t]
	\begin{center}
		\includegraphics[width=0.45\textwidth]{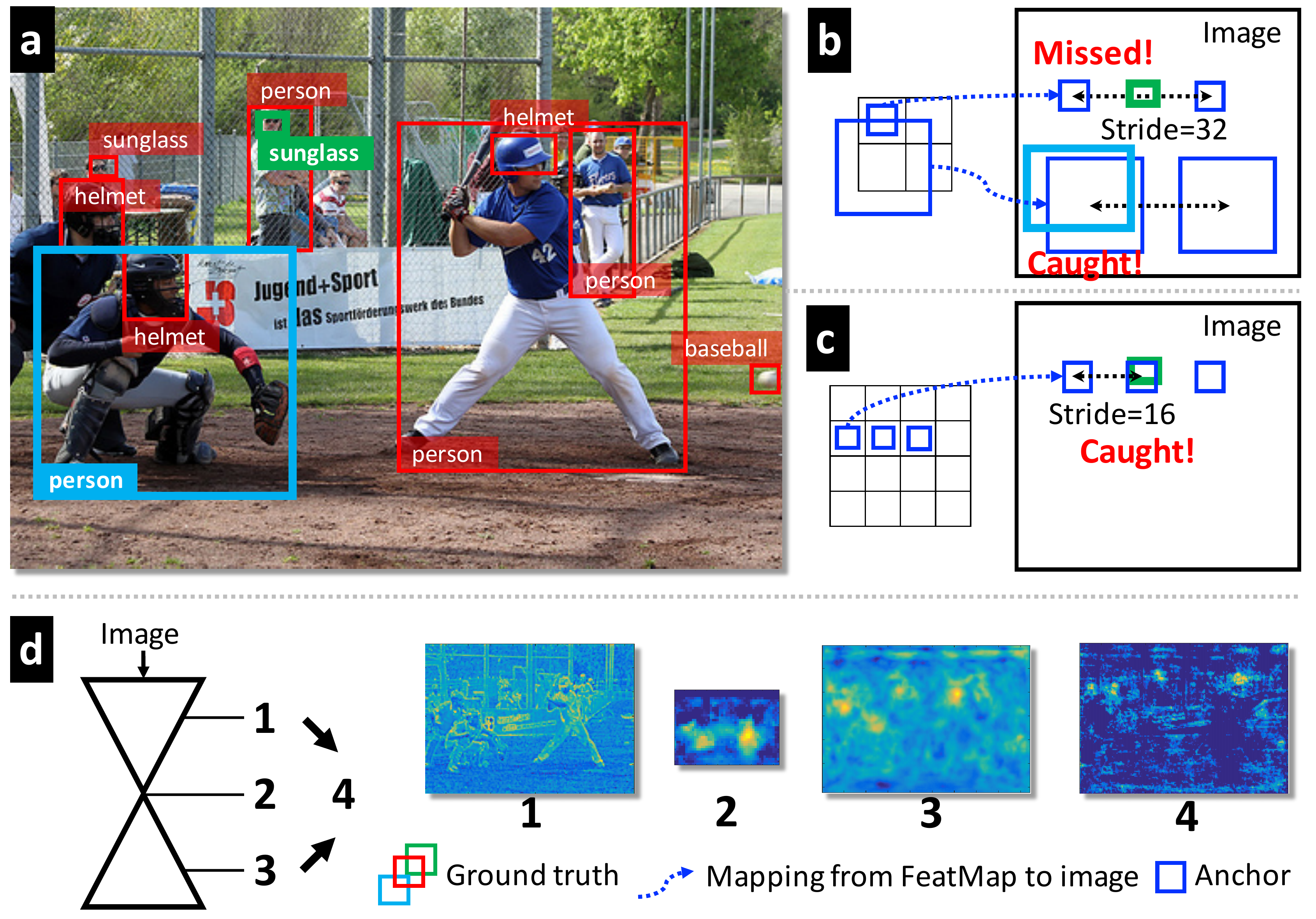}
	\end{center}
	\vspace{-.3cm}
	\caption{Motivation on using features of sufficient resolution. 
		{(a)} Input image where the sunglass in green and person in blue are being examined.
		{(b)} Smaller anchors fail to catch the sunglass when placed at the final map with stride 32. 
		{(c)} Smaller anchors succeed in catching the sunglass when placed at feature map with stride 16. 
		{(d)} Visualization of different map outputs in a zoom-in-and-out network.
	}
	\label{fig:motivation}
	\vspace{-.3cm}
\end{figure}

Object proposal is the task of proposing a set of candidate regions or bounding boxes in an image that may potentially contain an object.
In recent years, the emergence of object proposal algorithms \cite{selective_search,prime,MCG,craft,co_generate,hyper_net,deep_proposal,object_proposal_eval,pronet}
have significantly boosted the development of many vision tasks \cite{li2016cnn_sal,li2016multi,czz_tracking}, especially for  object detection \cite{rcnn,rfcn,fast_rcnn,inside_outside,ssd}.
It is verified by Hosang \textit{et.al} \cite{Hosang2015Pami} that region proposals with high average recall correlates well with good performance of a detector. Thus generating object proposals has quickly become the de-facto pre-processing step.


Currently, CNN models are known to be effective in generating candidate boxes \cite{faster_rcnn,deep_box,hyper_net}. Existing works use deep CNN features at the last layer for classifying whether a candidate box should be an object proposal. The candidate box can come from random seed \cite{attractioNet}, external boxes (selective search \cite{selective_search}, edge box \cite{edgebox}, \textit{etc.}), or 
sliding windows \cite{overfeat}. Deep CNN-based proposal methods employ a zoom-out network, in which sub-sampling is used for reducing the resolution of features. 
This zoom-out design is good for image classification since 
sub-sampling is effective for achieving translation invariance, increasing the receptive field of features, and saving computation.

\begin{figure*}
	\begin{center}
		\includegraphics[width=0.8\linewidth]{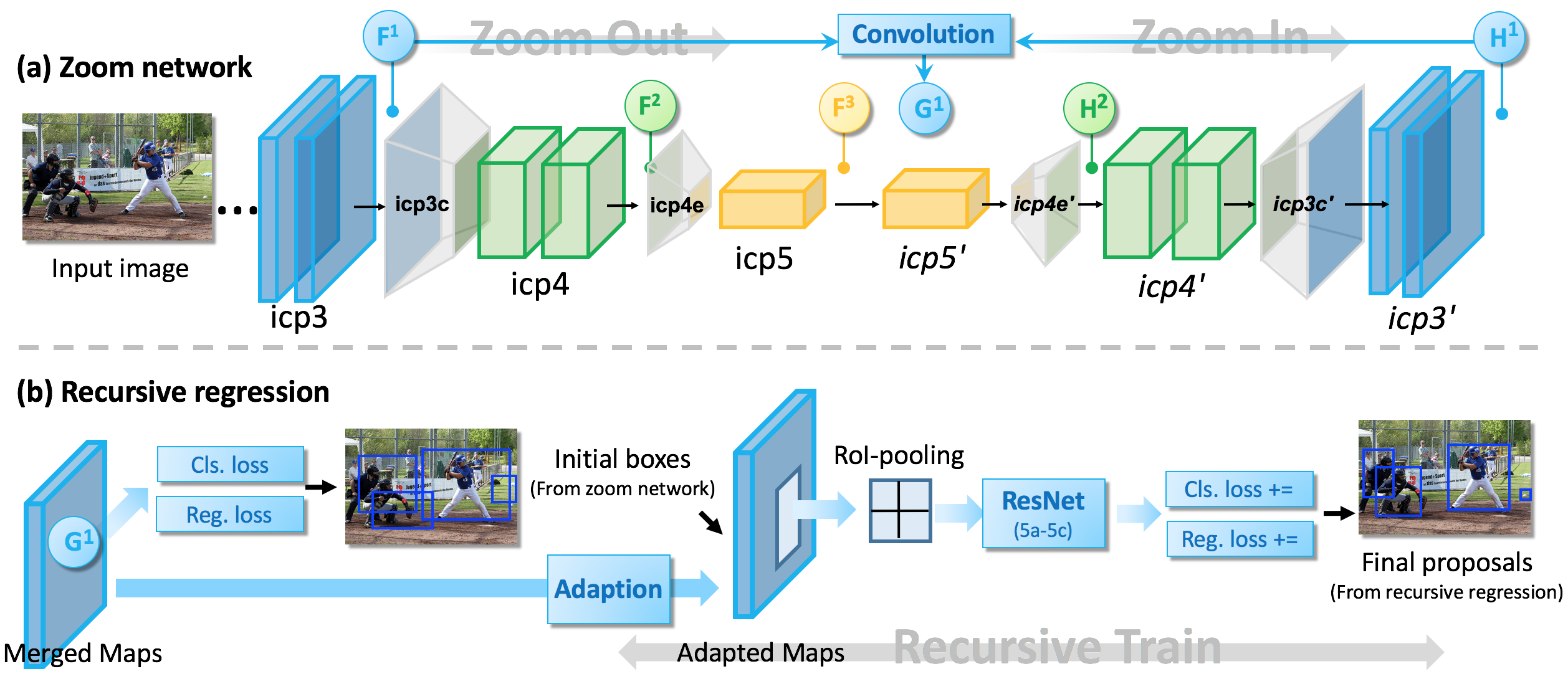}
	\end{center}
	\vspace{-.3cm}
		\caption{ZIP pipeline (only level $m=1$ are shown for brevity). 
			(a) Zoom out-and-in network. Each block stands for an inception module. Features $\mathbf{F}^1$ and $\mathbf{H}^1$ are combined into $\mathbf{G}^1$, which is responsible for detecting small objects
			(similarly for $\mathbf{G}^2$). Note that high-level maps alone are 
			suitable for detecting large objects due to 
			a larger stride and 
			the summarized semantics contained in the map
			(\textit{i.e.}, $\mathbf{F}^3 = \mathbf{G}^3$).
			(b) Regression and classification sub-network for obtaining object proposal.
			For recursive regression, residual blocks with RoI pooling are appended. 
		}
	\label{fig:network}
\end{figure*}

However, the zoom-out structure has two problems in object proposal for small objects.
%
First, candidate boxes, also called anchors, are placed on the final feature map in existing works. 
For instance, as shown in Figure~\ref{fig:motivation}(b),
when the down-sampling rate of the feature map is 32, moving the anchor by one step on the feature map corresponds to stepping by 32 pixels on the feature map.  
Due to this large stride on the image, anchors might skip small objects. 
%
Second, sub-sampling is good for translation invariance. However, it also makes it difficult to determine 
whether the feature is sub-sampled from foreground or background for small objects.
The lack of resolution in features is a factor that influences the ability of object proposal methods in finding small objects. 
As shown in Figure~\ref{fig:motivation}(c), if the resolution of feature map is sufficient, the anchor box with smaller stride can locate the small object. Simply increasing the resolution of image quadratically increases computation and memory, which is also limited by the current GPU memory size when a network goes deeper.
Lastly, objects of different sizes are recognized with features of the same resolution, which is inappropriate.

To fit for the zoom-out network design, one could 
resort to features at shallow layers with higher resolution. 
Features from shallow layers, however, are weak in extracting high-level information that is essential for object proposal. 
It is verified that higher layers often carry summarized high-level semantics \cite{visualize_cnn,alexnet}.
It would be good if the high-level sementics, with sufficient resolution, can be used for finding small objects.
Inspired by the hourglass network \cite{hg}, we propose a  zoom-out-and-in network for object proposal, namely,   \textbf{ZIP}.
Figure \ref{fig:network} illustrates the pipeline of our algorithm.
The zoom network gradually
upsamples high-level feature maps to sufficient resolution.
It classifies candidate region of different sizes with features of their corresponding resolutions.
Such 
%
a design takes full advantage of both low-level details and high-level semantics, which is illustrated in
Figure \ref{fig:motivation}(d) in a nutshell.
Feature 1 at shallower layer has more details about the image but cannot distinguish background from foreground. 
Feature 2 at deeper layer carries more semantic information, but is weak in locating small objects. 
If combined with Feature 3, which is deconvolved from Feature 2, the resultant Feature 4 is more useful in identifying small objects.

Iterative bounding box regression at the testing stage is found to be effective \cite{attractioNet,locNet,mrcnn}. However, the training stage is not aware of such iterative process. Therefore, there is a mismatch in the training and testing environment for regressing the targets of bounding boxes.
Gidaris \textit{et.al} \cite{attractioNet} create training samples in different regression stages by using a learned regressor in the intermediate stage or using an ideal label. However, these training samples still do not provide the learning algorithm with the actual setup at the testing stage. 
In this paper, we propose a new recursive training scheme 
where the regression targets are learned and refined consecutively during training. In this way, the iterative regression at the training stage is exactly the same as that in the testing stage.
Our regressor  is trained at one time and the network learns how to iteratively regress the bounding box to the ground-truth.
%
\smallskip

To sum up, our contributions in this work are as follows:
\begin{enumerate}
	\vspace{-.2cm}
	\item A zoom-out-and-in network that utilizes 
	feature maps in different semantic levels and resolutions to detect objects of various sizes.  
	Anchors are divided into clusters based on different scales and placed at feature maps with distinct strides.
	High-level semantics 
	are gradually 
	deconvolved and combined with 
	low-level, high-resolution maps to help identify small objects.
	
	\item A recursive training strategy to  regress bounding box locations iteratively, 
	where the setup of iterative regression in the training stage is consistent with that in the testing stage.
	
	\item The proposed ZIP algorithm achieves average recall to 61.3\% and 63.7\% at top 500 proposals on ILSVRC DET and MS COCO, respectively. 
	Furthermore, the proposed boxes will improve AP by around 2\% for object detection compared to previous state-of-the-art.
\end{enumerate}
%

ZIP is implemented in the Caffe \cite{caffe} toolkit with Matlab wrapper and trained on multiple GPUs. 
The codebase and object proposal results of our method are available\footnote{ \href{https://github.com/hli2020/zoom_network}{
	 \texttt{https://github.com/hli2020/zoom\_network}}}.

\vspace{-.1cm}
\section{Related Work}

The spirit of upsampling feature maps through learnable parameters is known as deconvolution, which is widely applied in different domains \cite{long2014fully,noh_iccv15, hypercolumn, hg}. 
Shelhamer \textit{et. al} \cite{long2014fully} first put forward a novel structure to do pixel-wise semantic segmentation via learnable deconvolution.
Recently, Newell \textit{et. al} \cite{hg} proposed an 
hourglass structure where  
feature maps are zoomed in and out through stacks for pose estimation.
%
Our model is inspired by these works and yet have distinctions in several ways:  
First, to the best of our knowledge, our work is the first in using the decovlution structure for object proposal.
Second, existing approaches either concatenate all features \cite{hypercolumn} or use the final feature map for prediction \cite{hg}, while we carefully design a network to select features at different locations (thus resolution varies) of a network for objects, \eg, low resolution features for large objects. 
With such a philosophy in mind, we have each object equipped with suitable features at a proper resolution.

Researchers are aware of the benefit of using features from different resolution levels. 
Jie \textit{et al.} \cite{scale_aware} proposed a scale-aware pixel-wise proposal framework 
where two separate networks are learned to handle large and small objects, respectively.
Yang \textit{et al.} \cite{scale_dependent_pooling} introduced a scale-dependent pooling scheme and exploited appropriate features depending on
the scale of candidate  proposals.
Recently, Liu \textit{et.al} \cite{ssd} proposed a fast end-to-end learning detector 
where different feature layers from the VGG model output individual predictions. Howerver, these works still use the zoom-out network structure and have limitations stated in Section \ref{Sec:intro}.

Bounding box regression is widely used in object detection and region proposal \cite{rcnn,fast_rcnn,faster_rcnn,spp}. Iterative bounding box regression is recently proved to be effective at the testing stage to improve the localization accuracy in many works \cite{locNet, attractioNet,mrcnn}. 
The bounding boxes approach closer to real locations step by step during iterative testing \cite{mrcnn}.
Our work in bounding box regression involves both training  and testing, which is not fully investigated in previous work.

\vspace{-.1cm}
\section{ZIP Algorithm}\label{sec:sop-algorithm}

\subsection{Network Architecture}\label{sec:network-architecture}

Figure \ref{fig:network} describes the overview of the proposed zoom out-and-in network, which contains three sub-networks.

\textbf{Zoom-Out Sub-network.} Most existing network structures \cite{alexnet,resNet,vgg} can be viewed and used as a zoom-out network. 
We adopt Inception-BN \cite{bn} throughout the paper. Specifically, 
an image is first fed into three convolutional layers, after which the feature maps are downsampled by a rate of 8. There are nine inception modules afterwards, denoted as 
\texttt{icp\_3a-3c}, \texttt{icp\_4a-4e} and \texttt{icp\_5a-5b}. Max-pooling is placed after \texttt{icp\_3c} and \texttt{icp\_4e}.

\textbf{Zoom-In Sub-network.}
Inspired by the hourglass network \cite{hg}, we adopt a zoom-in  architecture to better 
leverage the summarized high-level feature maps for refining its low-level counterparts.
%
The zoom-in architecture is exactly the mirrored version of the zoom-out network with max-pooling being replaced by deconvolution.
We denote 
\texttt{icp\_x'} as the mirrored version of its counterpart \texttt{icp\_x}. 


In this paper, we split the set of anchor candidates $\mathcal{A}$ into three clusters:
\begin{equation}
	\mathcal{A}=\{\mathcal{A}^m\}, m=1,\dots,3.
\end{equation}
The scales of anchors in each level are $\{16, 32\}$, $\{64, 128\}$, $\{256, 512\}$, respectively.
As is illustrated at the top of Figure \ref{fig:network}, we denote  
the feature maps from \texttt{icp\_3b}, \texttt{icp\_4d} and \texttt{icp\_5b}, 
as $\mathbf{F}^1$, $\mathbf{F}^2$ and $\mathbf{F}^3$;
%
their mirrored counterparts from
\texttt{icp\_3b'}, \texttt{icp\_4d'} and \texttt{icp\_5b'} as
$\mathbf{H}^1$, $\mathbf{H}^2$, and $\mathbf{H}^3$.
The \textit{combination} of feature maps $\mathbf{F}^m$ and $ \mathbf{H}^m$ on level $m$
are implemented via a convolution operation
:
\begin{align}
{\textbf{G}}^m  = \sigma(  \textbf{w}_F^m \otimes \textbf{F}^m  +  \textbf{w}_H^m \otimes \textbf{H}^m +  \textbf{b}^m),
\label{eq:message}
\end{align} 
where $\textbf{G}^m$ are the merged feature maps, $\otimes$ denotes the convolution, $\textbf{w}_H^m$ and $\textbf{w}_F^m$ are the filter weights, and $\textbf{b}^m$ represents the bias term. 

%

\begin{figure}
	\begin{center}
		\includegraphics[width=.45\textwidth]{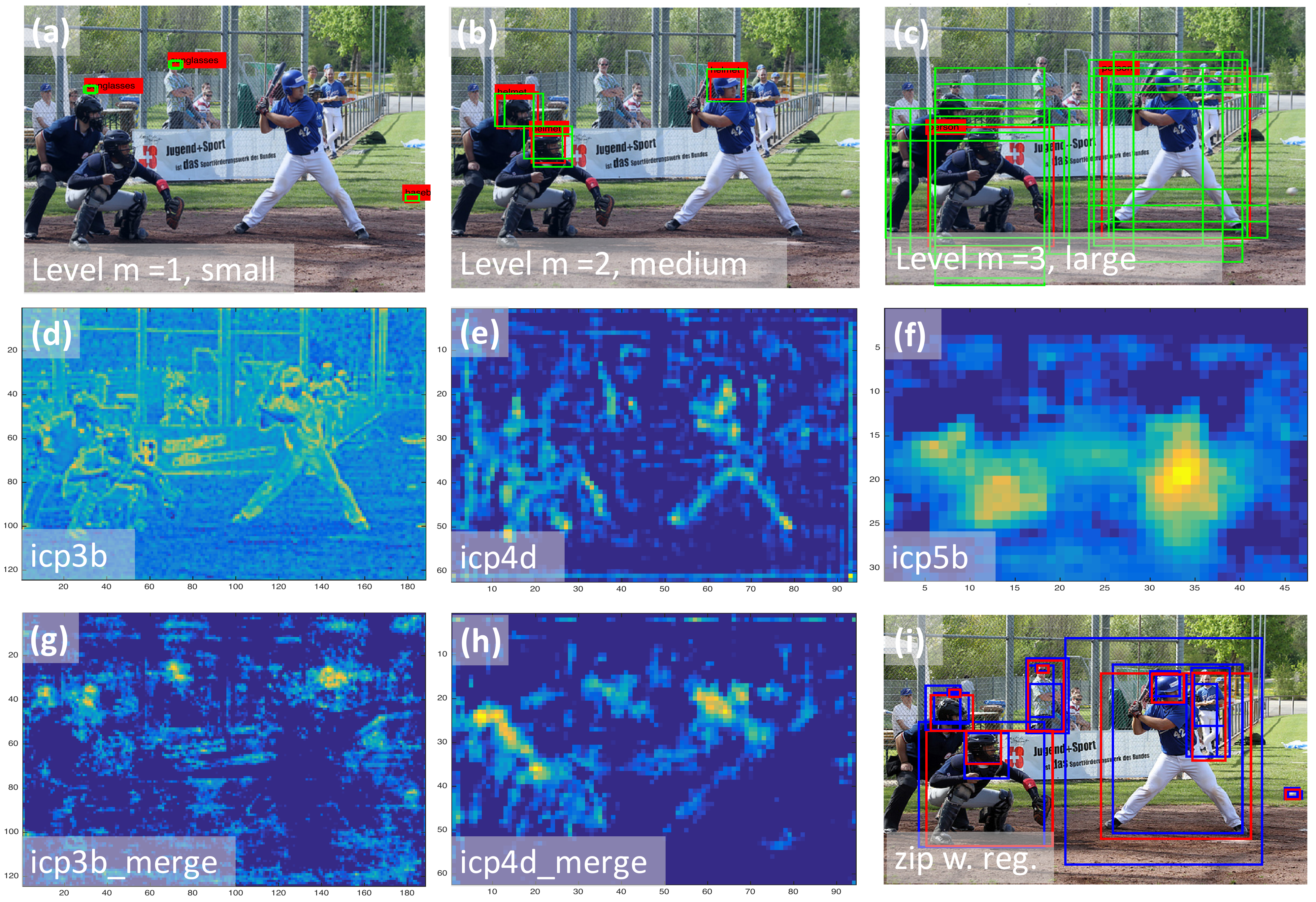}
	\end{center}
	\vspace{-.3cm}
	\caption{Anchor allocation and feature map breakdown. Ground truth are shown in red boxes.
		(a)-(c): anchor candidates with various scales are split and placed on feature maps with three resolutions (green boxes). 
		(d)-(h): feature maps at different layers in the network.
		(i): proposal results (blue boxes) after the RoI recursive regression network.
		Note that we resize higher level maps to the same size of level $m=1$ for better illustration.
	}
	\label{fig:featMap}
	\vspace{-.3cm}
\end{figure}

Figure \ref{fig:featMap} visualizes feature maps at different layers in the network.
$\mathbf{F}^1$(d) has high resolution and provides many local details when compared with $\mathbf{F}^3$(f). However, since $\mathbf{F}^1$ is shallow, it cannot distinguish object features from non-object features. This results in the difficulty in using $\mathbf{F}^1$ for the object-background classification. 
The upsampled feature map $\mathbf{H}^1$ after deconvolution is blessed with 
high-level semantic information for distinguishing foreground from background. 
After combining high-level semantic features from 
low-level features with local details, 
the resulting feature maps $\mathbf{G}^1$(g) are more useful for distinguishing objects from background. Similar analysis is applied to Figure \ref{fig:featMap}(e) and (h) when we identify the medium-sized helmets.

%

\textbf{Regression and Classification Sub-network.} 
The combined features $\mathbf{G}^m$ are served as input of the regression and classification sub-network. Each resolution level $m$ has two branches, as is shown at the bottom of Figure \ref{fig:network}.
In the first branch, $\mathbf{G}^m$ is fed into further classification and bounding box regression, \textit{i.e.}, generating object proposals from anchor set $\mathcal{A}^m$. 
After classification and regression 
in all resolution levels, 
all candidate boxes 
are merged and followed by a NMS with threshold 0.7. Roughly
2000 boxes with top confident scores, denoted by $\mathbf{R}$, are then used in the second RoI regression branch.

The input of the second branch is the combined feature maps $\mathbf{G}^m$ and the boxes $\mathbf{R}$ detected in the first branch. $\mathbf{R}$ is used by the RoI pooling layer to extract features for regions from $\mathbf{G}^m$. The features after RoI pooling are fed into three residual blocks \cite{resNet} to obtain features for recursive regression and classification.
%
%
After RoI pooling, the pooled features adapt to the aspect ratio and size of the box. This is different from the first branch, where boxes of different aspect ratios use the same $3\times 3$ feature map, and the used features are not adapted to aspect ratio or size change.

\subsection{Training with Recursive Regression}\label{sec:training-with-recursive-regression}
Let $L^m(p_i, t_i, k^*_i,  t^*_i)$ be the loss for sample $i$ on resolution level $m$, where $p_i=\{p_{i, k}|k=0,\ldots K\}$ 
is the estimated probability, 
$t_i$ indicates the estimated regression offset,  $k^{*}_i$ denotes the ground-truth class label, and $t^{*}_i$ represents the ground-truth regression offset.
The network is trained using the following loss function over the sample set $(\textbf{p}, \textbf{t}, \textbf{k}^*, \textbf{t}^*)$ on level $m$:
\begin{align}
L^m(\textbf{p}, \textbf{t}, \textbf{k}^{*},  \textbf{t}^{*}) &\\ 
 = -\frac{1}{N_{1}^m}   \sum_i & \log p_{i, k^{*}_i} +  \frac{1}{N_{2}^m} \sum_i [k^{*}_i=1] \mathcal{S}(t_i^{*}, t_i) ,   \nonumber
\label{loss}
\end{align}
where $\mathcal{S}$ is the smoothed $L_1$ loss between ground truth target $t_i^*$ and predicted target $t_i$, which is defined in \cite{fast_rcnn}. 
$N_{1}^m$ and $N_{2}^m$ are the batch size of classification and regression, respectively.
Thus the total loss is defined as the summed loss across all levels:
\begin{equation}
L = \sum_{m=1}^{M} 
L^m(\textbf{p}, \textbf{t}, \textbf{k}^{*},  \textbf{t}^{*}),
\label{total_loss}
\end{equation}
where $M$ is the number of resolution levels and $N$ being the batch size. 
%

For the RoI recursive training, let the probability estimated in the $q$th iteration and level $m$ for sample $i$ be $p^m_i(q)$; 
define the estimated region location and regression offset as $R^m_i(q)$, $t^m_i(q)$, respectively,
then $R^m_i(q)$ is obtained from $R^m_i(q-1)$ and $t^m_i(q)$ as:
\begin{equation}
R^m_i(q)  =R^m_i(q-1)+ \phi \left( t^m_i(q) \right),
\label{eq:Reg1}
\end{equation}
where $\phi( t^m_i(q))$ is the paramter transformation function that changes 
the representation of $t^m_i(q)$ defined in \cite{fast_rcnn} to the one that can be used as the regression offset.
Feature $\mathbf{G}^m_i(q-1)$ is obtained by RoI-pooling the combined feature map $\mathbf{G}^m$ using region $R^m_i(q-1)$. 
The feature $\mathbf{G}^m_i(q-1)$  is fed into three residual blocks for obtaining the features for 
$R^m_i(q-1)$. We use the following formulation to represent the feature extraction from residual blocks:
\begin{equation}
\mathbf{F}_{res, i}^m(q-1)  = f(\mathbf{G}^m_i(q-1), \theta_{res}),
\end{equation}
where $f(\cdot)$ is the feature extraction using the residual block and $\theta_{res}$ denotes the parameters in the residual block.
The features $\mathbf{F}_{res, i}^m(q-1)$ are then used for classification and bounding box regression:
\begin{gather}
p_{i}(q) = f_c(\mathbf{F}_{res, i}^m(q-1), \theta_c), \\
t_{i}(q) = f_r(\mathbf{F}_{res, i}^m(q-1), \theta_r),  \label{eq:Reg2}
\end{gather} 
where $\theta_c$ and $\theta_r$ are parameters used for classification and box regression, respectively.
The process in (\ref{eq:Reg1})-(\ref{eq:Reg2}) is repeated for $Q$ times as $q$ iterates from 1 to $Q$ in both training and testing stages. 

%
%
With the RoI recursive training included, the total loss $L$ will be modified to:
\begin{equation}
L = \sum_{m=1}^{M} \sum_{i=1}^N \sum_{q=1}^Q L^m \bigg(p_i(q), t_i(q), k^{*}_i(q), t^{*}_i(q) \bigg),
\label{total_loss_roi}
\end{equation}
where 
$Q$ is the number of iterative regression during training. $Q=1$ means applying regression only once. 
The gradient for loss in different regression stages are accumulated in a mini-batch for back-propagation.

There are several remarks regarding the training of the zoom network:
\begin{itemize}
	\item Adjust image scale at the training stage: 
	each training image is resized to the extent where at least one of 
	the ground truth boxes 
	is covered by anchors from $\mathcal{A}^m$. 
	This ensures there are always positive samples in each batch. 
	\item Control the number of negative samples in a batch. In preliminary experiments, 
	we find  training loss of the zoom network converges slowly if we fill up the rest of a batch with negative samples. 
	This will 
	cause the unbalance of training data for different classes when the number of positive samples is small. 
	Thus, we strict the number of negative samples to be twice the number of positive ones.
	\item Additional gray category. We find adding an additional gray label ($k$=2) into training will better separate the positive from the negative. For the positive class with $k$=1, the IoU should be above 0.6; for the gray class with $k$=2, the IoU ranges from 0.35 to 0.55; and for the negative with $k$=0, the IoU is below 0.25.
	The number of gray samples is set to be half of the total number of positive and negative ones. 
\end{itemize}

\subsection{Prediction}\label{sec:prediction}


For prediction in the zoom network, we take an inner-level and inter-scale NMS \cite{objectness} scheme. 
Since the scale varies dynamically during training, we also forward the network in several scales, 
ranging from 1400 to 200 with an interval of 200. For a certain scale, 
we concatenate output boxes from all the levels and conduct an inner-level NMS with a threshold of 0.7;
then we merge results from all scales and perform an inter-scale NMS with a threshold of 0.5.

For prediction in the RoI regression network, we average the results of bounding box regression as well as scores across all
levels and scales. The box locations are replaced by the new regression results while the scores are updated by adding the regression scores.
Then we conduct NMS with a threshold of 0.7 to have the final output of our ZIP algorithm. 

\begin{figure*}
	\begin{center}
		\includegraphics[width=.506\textwidth]{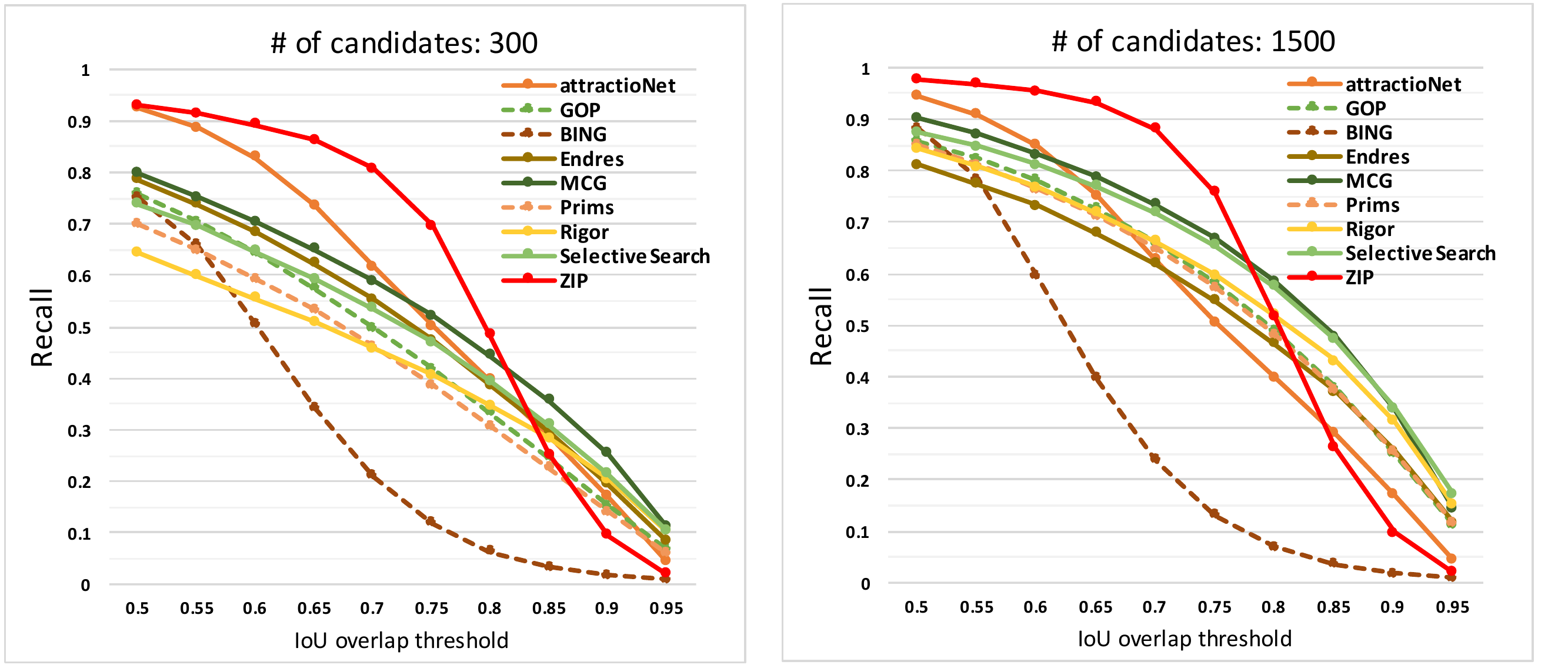}\includegraphics[width=.5\textwidth]{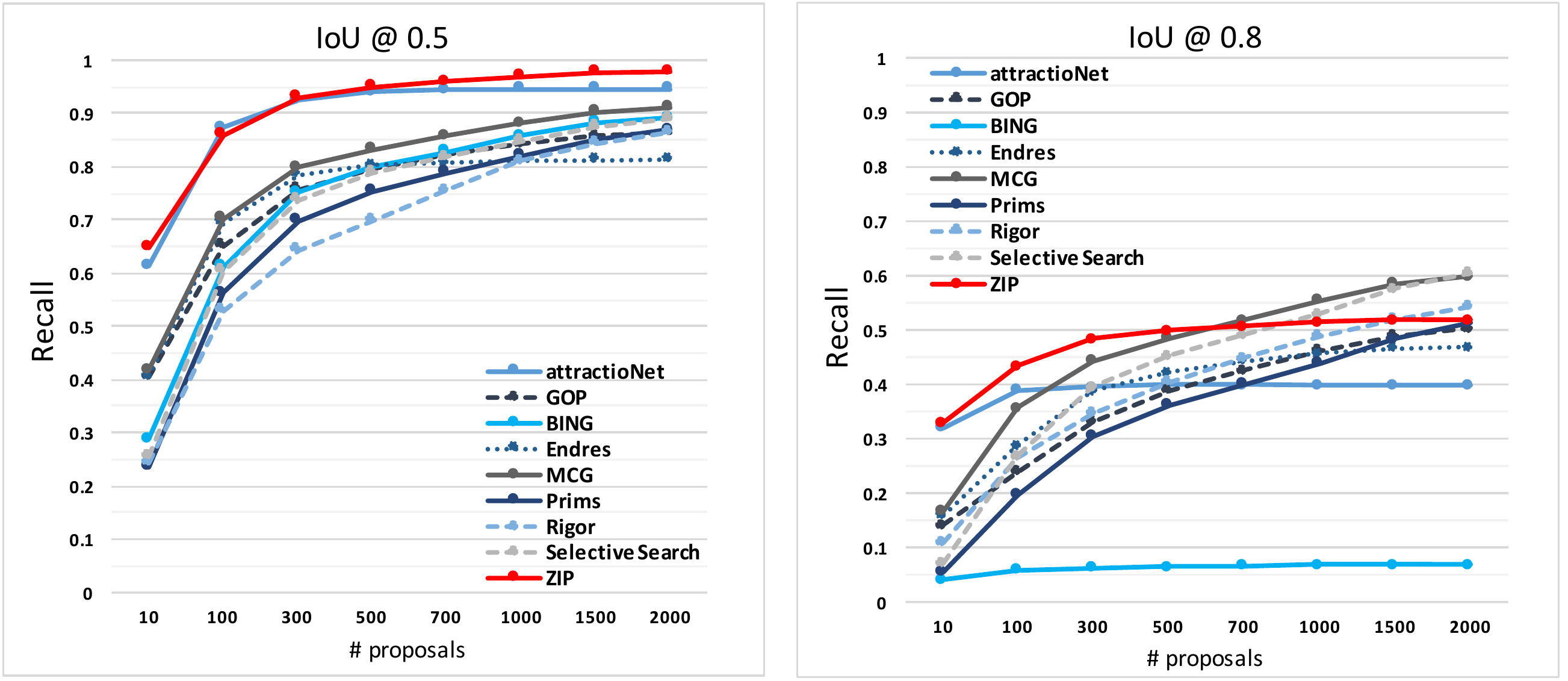}
		\includegraphics[width=.503\textwidth]{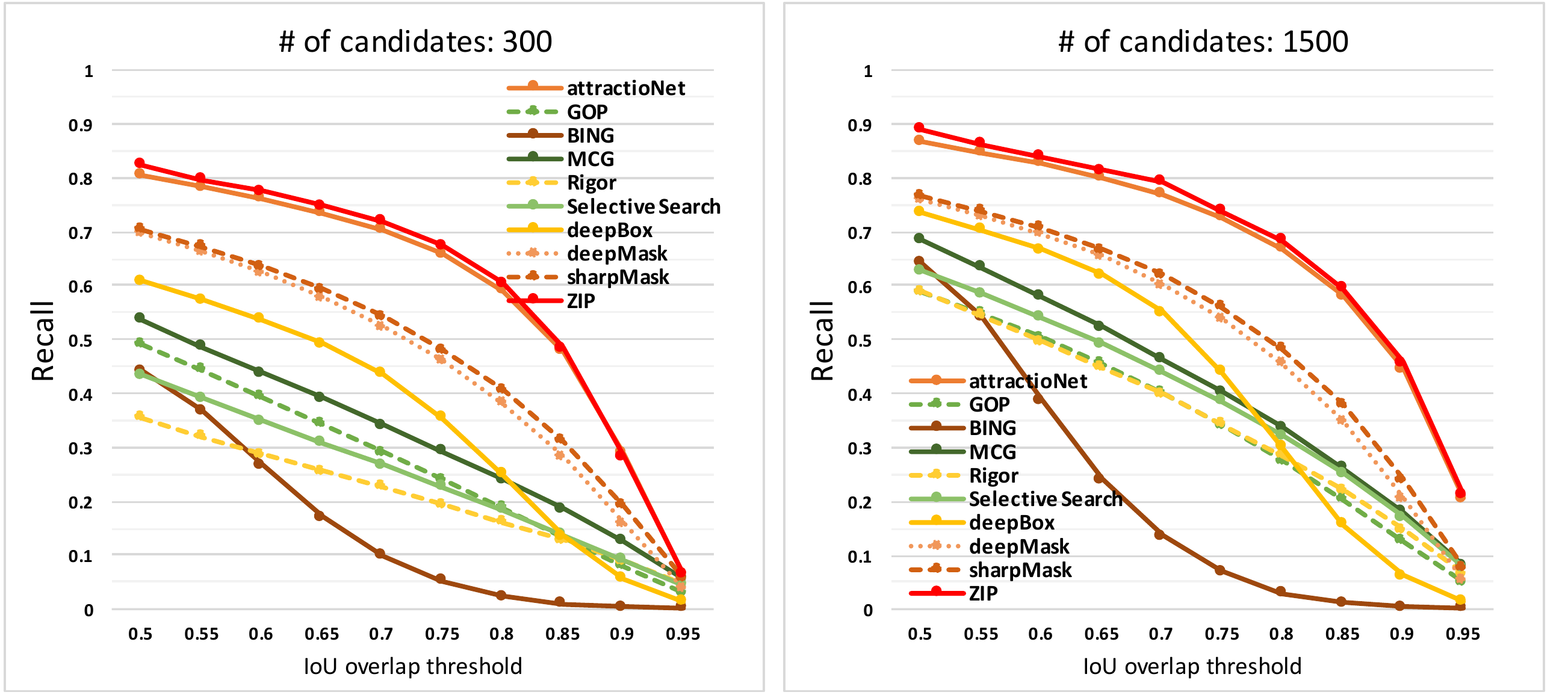}\includegraphics[width=.502\textwidth]{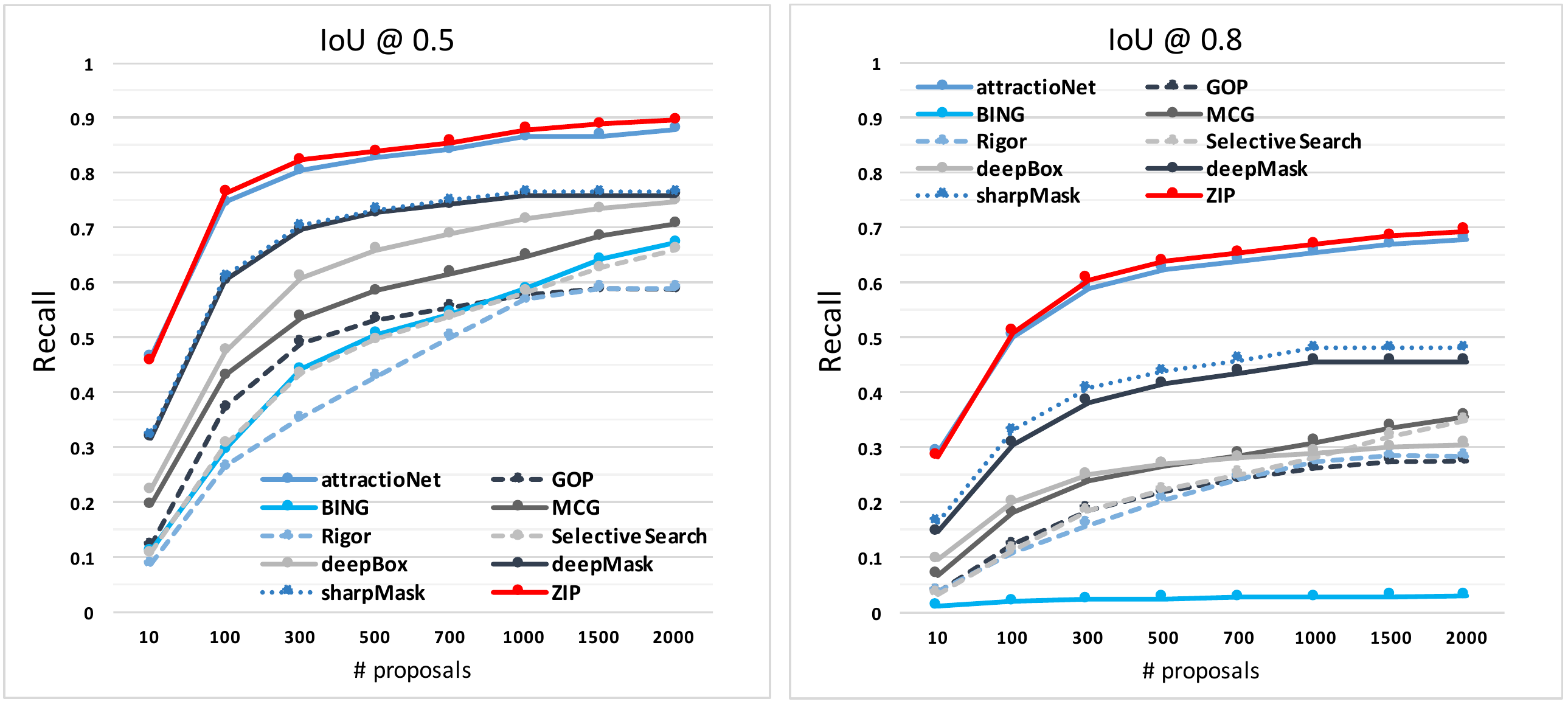}	
	\end{center}
	\vspace{-.4cm}
	\caption{Recall v.s. IoU thresholds (0.5, 0.8) and Number of Proposals (300, 1500), for ILSVRC DET 2014 ({top}) and MS COCO 2014 ({bottom}).
		Method abbreviations are provided in Table \ref{tab:grand_ilsvrc} and Table \ref{tab:grand_coco}.
	}
	\label{fig:recall_detail}
\end{figure*}

\section{Experiments}

\noindent In this section, we evaluate the effectiveness of our algorithm  on the problem of 
region proposal generation for generic objects. To this end, we evaluate our approach and compare our results with state-of-the-art object proposal methods on two
challenging 
datasets, ILSVRC DET 2014 \cite{imagenet_conf} and Microsoft COCO \cite{coco}.

\subsection{Datasets and setup}

The Microsoft COCO 2014 dataset \cite{coco} contains 82,783 training and 40,504 validation images, where most images have various shapes surrounded by complex scenes.
We use all the training images, without any data augmentation, to learn our model, and follow \cite{deepMask} in testing 5000 validation images for evaluation purpose (denoted as \texttt{val\_5k}).
The ILSVRC DET 2014 dataset \cite{imagenet_conf} is a subset of the whole ImageNet database and consists of more than 170,000 training and 20,000 validation images. Since some training images has only one object with simple background, which has a  distribution discretion with 
the validation set, we follow the practice of  \cite{fast_rcnn} and split the validation set into two parts. The training set is the \texttt{train\_14} with 44878 images and \texttt{val1} with 9205 images. 
We use \texttt{val2} as the validation set for evaluation. 
Flipping each image horizontally for training is used on ILSVRC but not used on COCO.

\textbf{Implementation details.} We pretrain an Inception-BN \cite{bn} on the ImageNet classification dataset,
which could achieve around 94\% top-5 accuracy in the 
classification task. Inception-BN is used as the zoom-out sub-network and its mirrored version is used as the zoom-in sub-network.
The three residual blocks are initialized from 
branch \texttt{5a} to \texttt{5c} of the ResNet model \cite{resNet} where we change stride from 2 to 1 using hole algorithm \cite{long2014fully}.
The base learning rate is set to 0.0001 with a 50\% drop every 7,000 iterations. The momentum and weight decay is set to be 0.9 and 0.0005, respectively. The maximum training iteration for both datasets is 200,000 (roughly 8 epochs) using four GPUs. The training process uses stochastic gradient descent optimization and takes around 1.5 days.
%
%
We use 
a batch size of 300 with each class having at most 100 samples for the first branch and 
a batch size of 
36 for the second branch. The number of iterations in the recursive regression is 2 for both training and testing if not specified. 
30 anchors 
are used with scale increase from 16 to 512 exponentially  
and aspect ratio being [0.15, 0.5, 1, 2, 6.7] on ILSVRC DET and [0.11, 0.4, 1, 2, 4.3] on MS COCO.
%
%

\textbf{Evaluation Metric.} We use \textit{recall} under different IoU thresholds and number of proposals as the main metric. The mean value of recall from IoU 0.5 to 0.95 is known as \textit{average recall} (AR). AR summarizes the general proposal performance and is shown to correlate  with the average precision (AP) performance of a detector better than other metrics \cite{Hosang2015Pami}.
Moreover,
we compute AR of different sizes of objects to further evaluate our algorithm's performance on a specific scale of targets. 
Following COCO-style evaluation, we denote three types of instance size, @small ($\alpha < 32^2 $), @medium ($ 32^2 \le \alpha < 96^2 $) and @large ($ 96^2 \le \alpha$), where $\alpha$ is the area of an object.

\subsection{ILSVRC DET 2014}

%
The top row of Figure~\ref{fig:recall_detail} illustrates recall@\texttt{prop}300, @\texttt{prop}1500 under different IoU overlap thresholds,
and recall@\texttt{IoU}0.5, @\texttt{IoU}0.8 under various number of region proposals.
Our algorithm is significantly better than others for IoU threshold from 0.6 to 0.8 and 
the number of proposals around 300. Table \ref{tab:grand_ilsvrc} 
reports the average recall v.s. the number of proposals (from 10 to 1000) and the size of objects\footnote{ We do not use the results directly from  \cite{attractioNet} since they tuned
optimal NMS thresholds for different cases. Instead, 
we fix the
NMS threshold as 0.5 for all the numbers of proposals through Table \ref{tab:grand_ilsvrc} and \ref{tab:grand_coco}. }.
Our method 
performs better on most metrics, especially for AR@small, where we improve by around 6\% compared with \cite{attractioNet}.
This 
proves that our pipeline is effective in detecting small objects.

\begin{table*}
	\begin{center}
		\footnotesize{
			\begin{tabular}{l c c c c c c c}
				\toprule
				\textbf{ILSVRC DET 2014} & AR@10 & AR@100 & AR@500 & AR@1000 & AR@Small & AR@Medium & AR@Large \\
				\midrule
				BING \cite{bing}  & 0.114 & 0.226 & 0.287& 0.307 & 0.000 & 0.064 & 0.340 \\
				EdgeBox \cite{edgebox} &  0.188 & 0.387 &  0.512 &0.555&0.021 &0.156 & 0.559  \\
				GOP \cite{gop} & 0.208 & 0.349 & 0.486 & 0.545 & 0.022 & 0.185 & 0.482 \\ 
				Selective Search \cite{selective_search} & 0.118 & 0.350 & 0.522 & 0.588 & 0.006 & 0.103 & 0.526\\
				MCG \cite{MCG} & 0.229 & 0.435 & 0.553 & 0.609 & 0.050 & 0.215 & 0.604   \\
				Endres \cite{cat_ind_obj_prop}  & 0.221 & 0.393 & 0.508 & 0.531 & 0.029 & 0.209 & 0.543  \\
				Prims \cite{prime} &  0.101 & 0.296 & 0.456 & 0.523 & 0.006 & 0.077 & 0.449 \\
				Rigor \cite{rigor} & 0.139 & 0.325 & 0.463 &0.551 & 0.027 & 0.092 & 0.485 \\
				Faster RCNN \cite{faster_rcnn} & 
				{}{0.356} & 0.475 & 0.532 & {}{0.560} & 0.217 & {}{0.407} & 0.571
				\\
				AttractioNet \cite{attractioNet} & {}{\textit{0.383}} & {}{\textit{0.514}} & {}{0.546} & 0.549 & {}{0.223} & {}{\textbf{0.449}} & {}{\textit{0.595}}  \vspace{.05cm} \\
				\midrule
				ZIP w/o regress & 0.247 & {}{0.483} & {}{\textit{0.601}} & {}{\textit{0.626}} & {}{\textit{0.232}} & {0.370} & {}{0.578} \\
				ZIP & {}{\textbf{0.401}} & {}{\textbf{0.539}} & {}{\textbf{0.613}} & {}{\textbf{0.631}} & 
				{}{\textbf{0.282}} & {}{\textit{0.448}} & {}{\textbf{0.626}} \\
				\bottomrule
			\end{tabular}
		}
	\end{center}
	\vspace{-.5cm}
	\caption{Average recall (AR) analysis on ILSVRC \texttt{val2}. The AR for small, medium and large objects are computed for 100 proposals. The top two results in each metric are in {\textbf{bold}} and {\textit{italic}}, respectively. 
		`w/o regress' means without regression.
	}
	\label{tab:grand_ilsvrc}
\end{table*}

Note that the performance of our proposed method drops when the number of proposals increase at IoU@0.8.
The threshold for the positive samples during training is 0.6,
meaning IoU@0.8 for test is harder to align due to a mismatch
with that during training; MCG \cite{MCG} or selective search \cite{selective_search} localizes objects
better when many boxes are generated, probably due to the
semantic merging on a superpixel level. Overall,
the drop corresponds to a large number of proposals (typically
larger than 800), and detectors will not use so many
proposals in practice (300-500 are enough, where ZIP is significantly
higher than others). Such a drop does not affect
detection performance.

\subsection{MS COCO}

For MS COCO, the bottom row of Figure \ref{fig:recall_detail} and Table \ref{tab:grand_coco} shows 
the same evaluation metrics as those on the ILSVRC dataset. 
Our ZIP algorithm still outperforms most state-of-the-arts in terms of recall and AR.

\begin{table*}
	\begin{center}
		\footnotesize{
			\begin{tabular}{l c c c c c c c}
				\toprule
				\textbf{MS COCO 2014} & AR@10 & AR@100 & AR@500 & AR@1000 & AR@Small & AR@Medium & AR@Large  \\
				\midrule
				BING \cite{bing}  & 0.042 & 0.100 & 0.164 & 0.189 & 0.000 & 0.026 & 0.269 \\
				EdgeBox \cite{edgebox} &  0.074 & 0.178 & 0.285 & 0.338 &0.009 & 0.086 &0.423\\
				GOP \cite{gop} & 0.058 & 0.187 & 0.297 & 0.339 & 0.007 & 0.141 & 0.401 \\
				Selective Search \cite{selective_search} & 0.052 & 0.163 & 0.287 & 0.351 & 0.003 &0.063 &0.407\\
				
				MCG \cite{MCG} &  0.098 & 0.240 & 0.342 & 0.387 & 0.036 & 0.173 & 0.497 \\
				
				Endres \cite{cat_ind_obj_prop} & 0.097 & 0.219 & 0.336 & 0.365 &0.013 & 0.164 &0.466\\
				DeepBox \cite{deep_box} & 0.127 & 0.270 & 0.376 &0.410 &0.043  & 0.239 & 0.511 \\
				CoGen$^{+}$ \cite{co_generate} &0.189 & 0.366 & -  & 0.492 & 0.107 & 0.449 & 0.686 \\
				DeepMask \cite{deepMask} & 0.183 & 0.367 & 0.470 & 0.504 & 0.065 & 0.454 & 0.555 \\
				SharpMask \cite{sharpMask} & {}{0.196} & 0.385 & 0.489& 0.524 & 0.068 & {}{0.472} &{}{0.587}\\
				AttractioNet \cite{attractioNet} &{}{\textbf{0.316}} & {}{\textit{0.519}} & {\textit{0.620}}& {}{\textit{0.651}} & {}{\textit{0.261}} & {}{\textbf{0.570}} & {}{\textit{0.705}}  \\
				\midrule
				ZIP {w/o regress} & 0.195 & {}{0.398} & {}{0.506} & {}{0.536} & {}{0.222} & 0.385 & 0.544\\
				ZIP &   {}{\textit{0.312}} & {}{\textbf{0.524}} & {}{\textbf{0.637}} & {}{\textbf{0.663}} & {}{\textbf{0.282}} & {}{\textit{0.564}} & {}{\textbf{0.709}} \\
				\bottomrule
			\end{tabular}
		}
	\end{center}
	\vspace{-.5cm}
	\caption{Average recall (AR) analysis on COCO \texttt{val\_5k}. The AR for small, medium and large objects are computed for 100 proposals. The top two results in each metric are in {\textbf{bold}} and {\textit{italic}}, respectively. 
		 $^{+}$: Results directly cited from \cite{co_generate}. 
		`w/o regress' means without regression.
	}
	\label{tab:grand_coco}
\end{table*}

\subsection{Ablation study}\label{sec:ablation-study}

\begin{table}
	\begin{center}
		\footnotesize{
			\begin{tabular}{l c}
				\toprule
				Structure & Rec@0.5  \\
				\midrule
				Zoom-out & 87.03 \\
				Zoom-out + \texttt{splitAnc} & 89.27 \\
				Deeper Zoom-out + \texttt{splitAnc} & 90.54 \\
				{Zoom Out-and-In} \small {Network} & 91.76  \\
				\bottomrule
			\end{tabular}
		}
	\end{center}
	\vspace{-.5cm}
	\caption{Ablation study on the zoom network structure. We use 30 anchors and treat training as a two-class problem. 
		}\label{tab:net_structure}
\end{table}

We investigate the effectiveness of different components in our algorithm through a series of ablation studies.
All the experiments 
use recall at IoU 0.5 (denoted as Rec@0.5) and AR of the top 300 proposals on the ILSVRC dataset. 

\textbf{Network design.} Table \ref{tab:net_structure} reports results for different network design of the zoom network. 
The second branch of the regression sub-network is not employed.
If we only use the zoom-out sub-network and place all anchor boxes at the last feature map, the recall is 87.03\%.
Then we split the anchors into three groups and use features from \texttt{icp\_3b}, \texttt{icp\_4d} and \texttt{icp\_5b} in the zoom-out sub-network as 
 features of for different-sized boxes.
 Such a  modification will increase  recall by around 2\%. 
Adding the zoom-out-and-in design further increases performance by around 2.5\%. 
The zoom-out-and-in network increases the depth of zoom-out network and has 40 layers. By simply stacking depth of the network to 40 layers using a zoom-out design (Deeper Zoom-out + \texttt{splitAnc} in Table \ref{tab:net_structure}, we do not witness an obvious increase (90.54) compared with the zoom-out-and-in structure.

\begin{table}
	\begin{center}
		\footnotesize{
		\begin{tabular}{l c c c}
			\toprule
			Scheme &  Train Rec. & Rec@0.5 & AR \\
			\midrule
			9 anchors (\textit{short for} ac.) 				& 82.02 & 86.41 & -  \\
			30 ac. 							& 84.31 & 89.45 & - \\
			30 ac. + \texttt{learnStats} 	& 88.24 & 91.76 & 47.54\\
			~~~~~~+ \texttt{dyTrainScale} & - & 0.88 $\uparrow$  & - \\
			~~~~~~+ \texttt{ctrlNegRatio} & - & 2.37 $\uparrow$ & -\\
			~~~~~~+ \texttt{grayCls} & - & 0.31 $\uparrow$  & -\\
		
			ZIP + 30 ac. + \texttt{all} & - & \textbf{94.63} &50.14 \\
			\bottomrule
		\end{tabular}
	}
	\end{center}
	\vspace{-.5cm}
	\caption{Ablation study on the anchor design and sampling scheme. 
		We use the zoom network in all settings. 
		`$\times \times \uparrow$' denotes absolute increase of recall by each individual strategy (see Section \ref{sec:ablation-study} for details)
		based on and compared to  the `30 ac. + \texttt{learnStats}' setting. `\texttt{all}' means adopting all the strategies together.
		}
	\label{tab:design}
\end{table}

\textbf{Anchor and training design.} 
Table \ref{tab:design} shows 
recall and AR on various anchor designs and sampling schemes. Only the first branch of the regression sub-network is used.
Using the original setting with 3 scales and 3 ratios as in \cite{faster_rcnn}, we have a 86.41\% recall@0.5.
By extending the number of scales to 6 and the number of aspect ratios to 5, anchor templates are increased to 30. 
If we linearly extend the aspect ratio to [0.25 0.5 1 2 4], recall is 89.45\%, denoted by {`30 ac.'} in Table \ref{tab:design}. 
After investigating the statistics of aspect ratio in the training data, we modify the set of aspect ratio to [0.15, 0.5, 1, 2, 6.7],
which is denoted as `30 ac. + \texttt{learnStats}' in Table \ref{tab:design}.
It is observed that such a data-driven setting of aspect ratio 
enhances recall by 4\% and 2.5\% in the training and validation set, respectively.
Furthermore,
we find that adjusting the scale of training image  (\texttt{dyTrainScale}),
the number of negative samples in a mini-batch (\texttt{ctrlNegRatio}), 
adding an additional gray class (\texttt{grayCls}) will increase recall on top of its previous setting.
After adopting all the aforementioned strategies, we have the {final non-regression version of ZIP} and achieve
a 94.63\% recall at IoU threshold 0.5 and an average recall of 50.14\%.

\begin{table}
	\begin{center}
		\footnotesize{
			\begin{tabular}{l  c c}
				\toprule
				Design &  Rec@0.5 & AR \\
				\midrule
				Simple & 93.01 & 56.30 \\
				ResNet + \texttt{oneStream} & 95.04 & 50.54 \\
				ResNet& 92.82 & 59.45 \\
				\rowcolor{lightgray} ResNet +  \texttt{grayCls} & \textbf{93.23} & \textbf{59.48}  \\
				\midrule
			    AtractioNet \cite{attractioNet}  & 92.45 & 53.93 \\						
				\bottomrule
			\end{tabular}
		}
	\end{center}
	\vspace{-.5cm}
	\caption{Ablation study on the design of the regression and classification sub-network structure. 
		The repetition number for training and test are all set to be $T=1$. See Section \ref{sec:ablation-study} for each structure's details.}
	\label{tab:roi_structure}
\end{table}

\begin{figure}[t]
	\begin{center}
		\includegraphics[width=0.35\textwidth]{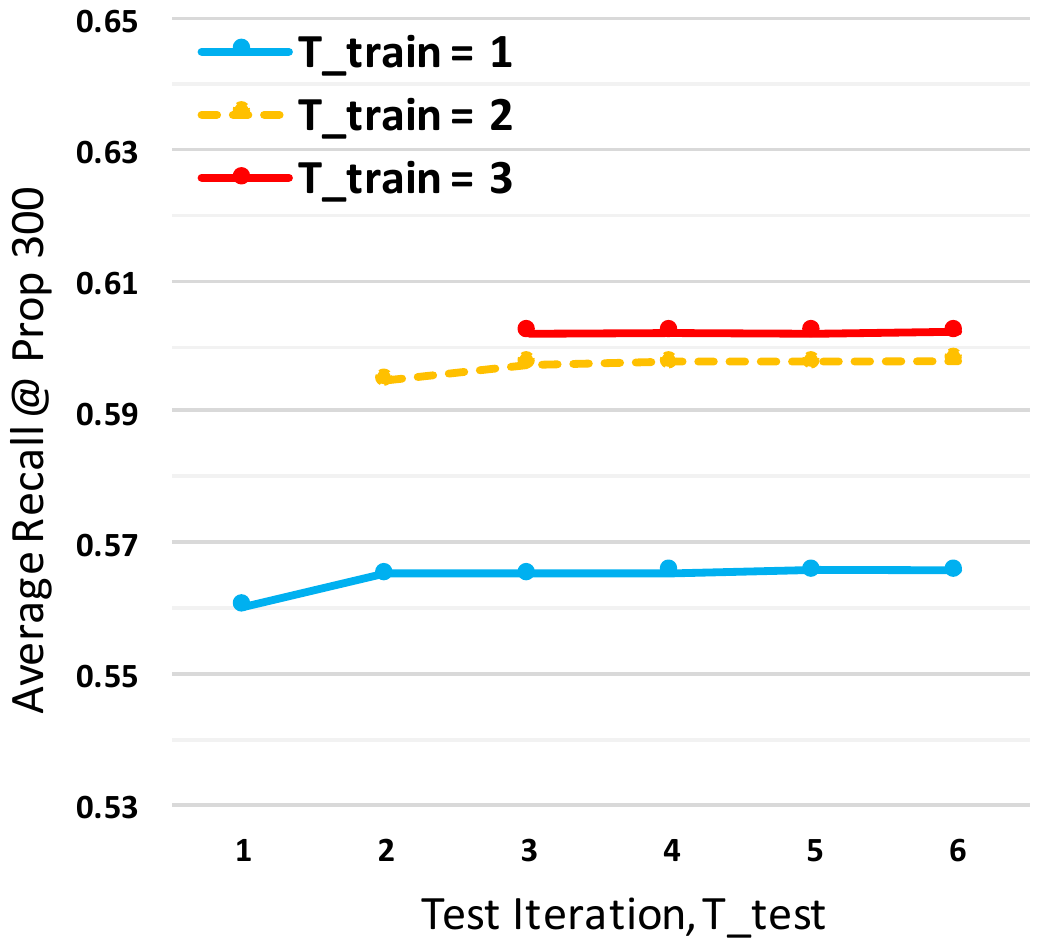}
	\end{center}
	\vspace{-.6cm}
	\caption{Investigation on the number of regression repetitions. 
		Note that $T_{train}=1$ or $T_{test}=1$ means applying box regression only once 
		without recursive training. 
		}
	\label{fig:recursive_regress}
\end{figure}

\textbf{Recursive regression design.}
Table \ref{tab:roi_structure} shows some variants of the regression sub-network structure with the second branch  included. 
%
The input boxes into the RoI regression network are from the best results in Table \ref{tab:design}, last row.
First, we append only one convolution layer after the RoI-pooling 
layer and before the global average pooling (known as {`simple'}). 
Note that boxes of different sizes will use their own features at the corresponding resolution. 
By using a recursive regression with 2 iterations, AR boosts from 50.14\% to 56.3\%. 
If we increase  depth in the recursive regression branch from one convolutional layer to three residual blocks ({`ResNet'}), 
AR boosts from 56.3\% to 59.45\%. If we simply concatenate features from different resolutions obtained from RoI-pooling, however, 
AR drops by about 9\%. Therefore, simple concatenation of features is worse than using features with suitable resolution.
We find that adding the additional gray class during training 
improves performance by 0.4\% (which is our final model).
Note that the additional regression step 
could significantly increase AR from 50.14 to 59.48; 
but decrease recall slightly. 
This is because object proposals from the recursive regression will move to more confident boxes but ignore some less confident ones that may contain an object. 

\textbf{Number of recursive regression.}
Figure \ref{fig:recursive_regress} investigates on the number of regression repetition in both training and test stage.
It is observed that  $T=2$ is found to be sufficient in reaching high AR. 
There is a big gain between using recursive training ($T=2$) and without ($T=1$).
More recursive iterations will slightly improve AR and yet may not be necessary 
considering algorithm's efficiency. Therefore, we adopt $T=2$ as the default setting.
%


To sum up, the final version of ZIP is the zoom out-and-in network with recursive regression $T=2$, which is shown in Table \ref{tab:roi_structure} with gray background 
and denoted as `ResNet + \texttt{grayCls}'.

\subsection{Evaluation in detection system}


 \begin{table}
 	\begin{center}
 		\footnotesize{
 			\begin{tabular}{l c c c }
 				\toprule
 				Method & @0.50 & @0.75 & @0.5:0.95 \\
 				\midrule
 				EdgeBox \cite{edgebox} & 49.94 & 35.24 & 31.47\\
 				AttractioNet \cite{attractioNet} & 50.19 & 33.45 & 32.26 \\
 				Selective Search \cite{selective_search} & 51.98 & 35.13 & 34.22 \\
 				ZIP & \textbf{53.92} & \textbf{35.59} & \textbf{34.37} \\
 				\midrule
 				Method & @small & @medium & @large \\
 				\midrule
 				AttractioNet \cite{attractioNet} & 2.06 & 17.81 & 54.83  \\
 				EdgeBox \cite{edgebox} & 3.98 & 20.61 & 60.98 \\
 				Selective Search \cite{selective_search} & 4.86 & 23.24 & 61.78 \\
 				ZIP &  \textbf{4.86} & \textbf{24.31} & \textbf{62.02} \\
 				\bottomrule
 			\end{tabular}
 		}
 	\end{center}
 	\vspace{-.5cm}
 	\caption{Detection results on ILSVRC DET 2014: AP performance at different IoUs 
 		using R-FCN detector \cite{rfcn}
 		with top 300 boxes.
 		}
 	\label{tab:detection}
 \end{table}

At last we evaluate the proposed algorithm in the detection system.
We follow the same training and test procedure as that in the R-FCN detector \cite{rfcn}.
The evaluation metric is the average precision (AP) at different IoU thresholds as well as a COCO-style AP that averages IoU thresholds from 0.5 to 0.95.
Table \ref{tab:detection} reports the detection performance of different proposal methods using top 300 
boxes. 
It can be seen that our object proposal provides a better mAP 
on the ILSVRC DET dataset and are suitable for detecting objects of different sizes.
%
\section{Conclusion}

In this work, we have proposed a zoom-out-and-in network 
that utilizes both low-level details and high-level semantics. 
The 
information from top layers is gradually upsampled by deconvolution to reach suitable resolution for small-sized objects.  
The zoom-in-and-out pipeline employs features from different resolutions in a network to
detect objects of various sizes. Such a strategy could alleviate the drawback of identifying small objects on feature maps with a large stride.
We further propose a recursive training scheme to do multiple iterations of regression to better refine the bounding boxes, yielding a higher average recall on both ILSVRC DET and MS COCO datasets. Region proposals from the zoom network is found to provide around 2\% mean AP gain for object detection when compared with other state-of-the-art object proposals.

\section*{Acknowledgement}
We would like to thank S. Gidaris, X. Tong and K. Kang for helpful discussions along the way, W. Yang for proofreading the manuscript.  H. Li is funded by the Hong Kong Ph.D Fellowship scheme. We are also grateful for SenseTime Group Ltd. donating the resource of GPUs.

{\footnotesize
\bibliographystyle{ieee}
\bibliography{deep_learning,obj_prop}

\begin{thebibliography}{10}\itemsep=-1pt

\bibitem{objectness}
B.~Alexe, T.~Deselaers, and V.~Ferrari.
\newblock Measuring the objectness of image windows.
\newblock {\em IEEE Trans. Pattern Anal. Mach. Intell.}, 34(11):2189--2202,
  Nov. 2012.

\bibitem{MCG}
P.~Arbel\'{a}ez, J.~Pont-Tuset, J.~Barron, F.~Marques, and J.~Malik.
\newblock Multiscale combinatorial grouping.
\newblock In {\em CVPR}, 2014.

\bibitem{inside_outside}
S.~Bell, C.~L. Zitnick, K.~Bala, and R.~Girshick.
\newblock Inside-outside net: Detecting objects in context with skip pooling
  and recurrent neural networks.
\newblock In {\em CVPR}, 2016.

\bibitem{object_proposal_eval}
N.~Chavali, H.~Agrawal, A.~Mahendru, and D.~Batra.
\newblock Object-proposal evaluation protocol is 'gameable'.
\newblock In {\em CVPR}, 2016.

\bibitem{bing}
M.~Cheng, Z.~Zhang, W.~Lin, and P.~H.~S. Torr.
\newblock {BING:} binarized normed gradients for objectness estimation at
  300fps.
\newblock In {\em CVPR}, 2014.

\bibitem{czz_tracking}
Z.~Chi, H.~Li, H.~Lu, and M.-H. Yang.
\newblock Dual deep network for visual tracking.
\newblock {\em arXiv preprint: 1612.06053}, 2016.

\bibitem{rfcn}
J.~Dai, Y.~Li, K.~He, and J.~Sun.
\newblock {{R-FCN}: Object Detection via Region-based Fully Convolutional
  Networks}.
\newblock In {\em NIPS}, 2016.

\bibitem{imagenet_conf}
J.~Deng, W.~Dong, R.~Socher, L.-J. Li, K.~Li, and L.~Fei-Fei.
\newblock {ImageNet: A Large-Scale Hierarchical Image Database}.
\newblock In {\em CVPR}, 2009.

\bibitem{cat_ind_obj_prop}
I.~Endres and D.~Hoiem.
\newblock Category-independent object proposals with diverse ranking.
\newblock {\em IEEE Trans. on PAMI}, 36:222--234, 2014.

\bibitem{deep_proposal}
A.~Ghodrati, A.~Diba, M.~Pedersoli, T.~Tuytelaars, and L.~V. Gool.
\newblock {DeepProposals}: Hunting objects and actions by cascading deep
  convolutional layers.
\newblock {\em arXiv preprint: 1606.04702}, 2016.

\bibitem{mrcnn}
S.~Gidaris and N.~Komodakis.
\newblock Object detection via a multi-region and semantic segmentation-aware
  cnn model.
\newblock In {\em CVPR}, 2015.

\bibitem{attractioNet}
S.~Gidaris and N.~Komodakis.
\newblock {Attend Refine Repeat}: Active box proposal generation via in-out
  localization.
\newblock In {\em BMVC}, 2016.

\bibitem{locNet}
S.~Gidaris and N.~Komodakis.
\newblock {LocNet}: Improving localization accuracy for object detection.
\newblock In {\em CVPR}, 2016.

\bibitem{fast_rcnn}
R.~Girshick.
\newblock {Fast R-CNN}.
\newblock In {\em ICCV}, 2015.

\bibitem{rcnn}
R.~Girshick, J.~Donahue, T.~Darrell, and J.~Malik.
\newblock Rich feature hierarchies for accurate object detection and semantic
  segmentation.
\newblock In {\em CVPR}, 2014.

\bibitem{hypercolumn}
B.~Hariharan, P.~Arbeláez, R.~Girshick, and J.~Malik.
\newblock Hypercolumns for object segmentation and fine-grained localization.
\newblock In {\em CVPR}, 2014.

\bibitem{co_generate}
Z.~Hayder, X.~He, and M.~Salzmann.
\newblock Learning to co-generate object proposals with a deep structured
  network.
\newblock In {\em CVPR}, 2016.

\bibitem{resNet}
K.~He, X.~Zhang, S.~Ren, and J.~Sun.
\newblock Deep residual learning for image recognition.
\newblock In {\em CVPR}, 2016.

\bibitem{Hosang2015Pami}
J.~Hosang, R.~Benenson, P.~Doll\'ar, and B.~Schiele.
\newblock What makes for effective detection proposals?
\newblock {\em IEEE Trans. on PAMI}, 2015.

\bibitem{rigor}
A.~Humayun, F.~Li, and J.~M. Rehg.
\newblock Rigor: Reusing inference in graph cuts for generating object regions.
\newblock In {\em CVPR}, 2014.

\bibitem{bn}
S.~Ioffe and C.~Szegedy.
\newblock Batch normalization: Accelerating deep network training by reducing
  internal covariate shift.
\newblock In {\em ICML}. 2015.

\bibitem{caffe}
Y.~Jia, E.~Shelhamer, J.~Donahue, S.~Karayev, J.~Long, R.~Girshick,
  S.~Guadarrama, and T.~Darrell.
\newblock Caffe: Convolutional architecture for fast feature embedding.
\newblock In {\em ACM Multimedia}, 2014.

\bibitem{scale_aware}
Z.~Jie, X.~Liang, J.~Feng, W.~F. Lu, E.~H.~F. Tay, and S.~Yan.
\newblock Scale-aware pixelwise object proposal networks.
\newblock {\em IEEE Trans. on Image Processing}, 25, 2016.

\bibitem{spp}
H.~Kaiming, Z.~Xiangyu, R.~Shaoqing, and J.~Sun.
\newblock Spatial pyramid pooling in deep convolutional networks for visual
  recognition.
\newblock In {\em ECCV}, 2014.

\bibitem{hyper_net}
T.~Kong, A.~Yao, Y.~Chen, and F.~Sun.
\newblock Hypernet: Towards accurate region proposal generation and joint
  object detection.
\newblock In {\em CVPR}, 2016.

\bibitem{gop}
P.~Krahenbuhl and V.~Koltun.
\newblock Geodesic object proposals.
\newblock In {\em ECCV}, 2014.

\bibitem{alexnet}
A.~Krizhevsky, I.~Sutskever, and G.~E. Hinton.
\newblock Imagenet classification with deep convolutional neural networks.
\newblock In {\em NIPS}, pages 1106--1114, 2012.

\bibitem{deep_box}
W.~Kuo, B.~Hariharan, and J.~Malik.
\newblock Deep{B}ox: Learning objectness with convolutional networks.
\newblock In {\em ICCV}, 2015.

\bibitem{li2016cnn_sal}
H.~Li, J.~Chen, H.~Lu, and Z.~Chi.
\newblock {CNN} for saliency detection with low-level feature integration.
\newblock {\em Neurocomputing}, 226:212--220, 2017.

\bibitem{li2016multi}
H.~Li, W.~Ouyang, and X.~Wang.
\newblock Multi-bias non-linear activation in deep neural networks.
\newblock In {\em ICML}, 2016.

\bibitem{coco}
T.-Y. Lin, M.~Maire, S.~Belongie, L.~Bourdev, R.~Girshick, J.~Hays, P.~Perona,
  D.~Ramanan, C.~L. Zitnick, and P.~Dollar.
\newblock {Microsoft COCO: Common Objects in Context}.
\newblock {\em arXiv preprint:1405.0312}, 2014.

\bibitem{ssd}
W.~Liu, D.~Anguelov, D.~Erhan, C.~Szegedy, and S.~Reed.
\newblock {SSD}: Single shot multibox detector.
\newblock In {\em ECCV}, 2016.

\bibitem{long2014fully}
J.~Long, E.~Shelhamer, and T.~Darrell.
\newblock Fully convolutional networks for semantic segmentation.
\newblock In {\em CVPR}, 2015.

\bibitem{prime}
S.~Man\'en, M.~Guillaumin, and L.~Van~Gool.
\newblock {Prime Object Proposals with Randomized Prim's Algorithm}.
\newblock In {\em ICCV}, 2013.

\bibitem{hg}
A.~Newell, K.~Yang, and J.~Deng.
\newblock Stacked hourglass networks for human pose estimation.
\newblock In {\em ECCV}, 2016.

\bibitem{noh_iccv15}
H.~Noh, S.~Hong, and B.~Han.
\newblock Learning deconvolution network for semantic segmentation.
\newblock In {\em ICCV}, 2015.

\bibitem{deepMask}
P.~O. Pinheiro, R.~Collobert, and P.~Dollar.
\newblock Learning to segment object candidates.
\newblock In {\em NIPS}, 2015.

\bibitem{sharpMask}
P.~O. Pinheiro, T.-Y. Lin, R.~Collobert, and P.~Dollár.
\newblock Learning to refine object segments.
\newblock In {\em ECCV}, 2016.

\bibitem{faster_rcnn}
S.~Ren, K.~He, R.~Girshick, and J.~Sun.
\newblock {Faster {R-CNN}: Towards Real-Time Object Detection with Region
  Proposal Networks}.
\newblock In {\em NIPS}, 2015.

\bibitem{overfeat}
P.~Sermanet, D.~Eigen, X.~Zhang, M.~Mathieu, R.~Fergus, and Y.~LeCun.
\newblock Overfeat: Integrated recognition, localization and detection using
  convolutional networks.
\newblock In {\em ICLR}, 2014.

\bibitem{vgg}
K.~Simonyan and A.~Zisserman.
\newblock Very deep convolutional networks for large-scale image recognition.
\newblock In {\em International Conference on Learning Representations}, 2015.

\bibitem{pronet}
C.~Sun, M.~Paluri, R.~Collobert, R.~Nevatia, and L.~Bourdev.
\newblock {ProNet}: Learning to propose object-specific boxes for cascaded
  neural networks.
\newblock In {\em CVPR}, 2016.

\bibitem{selective_search}
J.~Uijlings, K.~van~de Sande, T.~Gevers, and A.~Smeulders.
\newblock Selective search for object recognition.
\newblock {\em International Journal of Computer Vision}, 2013.

\bibitem{craft}
B.~Yang, J.~Yan, Z.~Lei, and S.~Z. Li.
\newblock {CRAFT} objects from images.
\newblock In {\em CVPR}, 2016.

\bibitem{scale_dependent_pooling}
F.~Yang, W.~Choi, and Y.~Lin.
\newblock Exploit all the layers: Fast and accurate cnn object detector with
  scale dependent pooling and cascaded rejection classifiers.
\newblock In {\em CVPR}, 2016.

\bibitem{visualize_cnn}
M.~D. Zeiler and R.~Fergus.
\newblock Visualizing and understanding convolutional networks.
\newblock {\em arXiv preprint: 1311.2901}, 2013.

\bibitem{edgebox}
L.~Zitnick and P.~Dollar.
\newblock Edge {B}oxes: Locating object proposals from edges.
\newblock In {\em ECCV}, 2014.

\end{thebibliography}
}

\end{document}